\documentclass{article}
\usepackage{spconf,amsmath,graphicx}
\usepackage{amsfonts}
\usepackage{booktabs}
\usepackage{multirow}
\usepackage{bm}
\usepackage[linesnumbered, lined, ruled]{algorithm2e}
\usepackage{microtype}

\title{FDDWNet: a lightweight Convolutional Neural Network \\ for real-time semantic segmentation}
\name{Jia Liu$^{1}$, Quan Zhou$^{1, }$\sthanks{Corresponding author: Quan Zhou, quan.zhou@njupt.edu.cn. This work is partly supported by NSFC (No. 61876093, 61801242, 61701252, 61671253), and NSFJS (No. BK20181393).}, Yong Qiang$^{1}$, Bin Kang$^{2}$, Xiaofu Wu$^{1}$, and Baoyu Zheng$^{1}$}
\address{$^{1}$National Engineering Research Center of Communications and Networking, \\ Nanjing University of Posts \& Telecommunications, P.R. China.\\
$^{2}$School of Internet of Things, Nanjing University of Posts \& Telecommunications, P.R. China.\\}

\begin{document}
%
\maketitle
\begin{abstract}

This paper introduces a lightweight convolutional neural network, called \emph{FDDWNet}, for real-time accurate semantic segmentation. In contrast to recent advances of lightweight networks that prefer to utilize shallow structure, \emph{FDDWNet} makes an effort to design more deeper network architecture, while maintains faster inference speed and higher segmentation accuracy. Our network uses factorized dilated depth-wise separable convolutions (FDDWC) to learn feature representations from different scale receptive fields with fewer model parameters. Additionally, \emph{FDDWNet} has multiple branches of skipped connections to gather context cues from intermediate convolution layers. The experiments show that \emph{FDDWNet} only has 0.8M model size, while achieves 60 FPS running speed on a single RTX 2080Ti GPU with a $1024 \times 512$ input image. The comprehensive experiments demonstrate that our model achieves state-of-the-art results in terms of available speed and accuracy trade-off on CityScapes and CamVid datasets.

\end{abstract}
\section{Introduction}
\label{sec:intro}

Semantic segmentation plays a significant role in computer vision, and facilitates some real-world applications, such as augmented
reality, robotics, and self-driving. It aims to partition an input image into a serious of disjointed image regions, in which each one is associated with pre-defined semantic labels including stuff (e.g. sky, road, grass) and discrete objects (e.g. person, car, bicycle). Recently, convolutional neural networks (CNNs), especially fully convolutional networks (FCNs) \cite{long2017fully,zhao2017pyramid} and encoder-decoder networks (EDNs) \cite{Guosheng2017RefineNet,jun2019dual}, have become a primary trend for solving semantic segmentation problems. Although these advances achieve remarkable progress by designing deeper and larger networks (e.g., VGGNet \cite{long2017fully} and ResNet \cite{Guosheng2017RefineNet,jun2019dual}), they often consume a huge amounts of resources, which is not suitable for edge devices (e.g., cellphones, robots, and drones) that have limited memory constraint and reduced computational capabilities.

In order to adapt to real-world scenarios that require real-time prediction and decision, recent efforts prefer to build lightweight networks \cite{li2018constrained,wen2016learning,Rastegari2016xnor,courbariaux2016binarized,Howard2017mobile,zhang2018shuffle} with shallow architecture, which can be roughly classified into three categories: (1) \emph{Network compression based methods} \cite{li2018constrained,wen2016learning} remove redundancies of a pre-trained model to boost efficiency through pruning techniques. (2) \emph{Low bitwise based approaches} \cite{Rastegari2016xnor,courbariaux2016binarized} use quantized technique to enhance efficiency, where the learned model weights are represented by few bits instead of high precision floating points. Unlike compression based methods, these models usually do not change the network structure, yet often come at a price of poor segmentation performance. (3) \emph{Lightweight CNNs} \cite{Howard2017mobile,zhang2018shuffle} directly target on computationally cheap network to improve efficiency, where convolutional factorization is often used to reduce model size. For example, ShuffleNets \cite{zhang2018shuffle,ma2018shufflenet} and MobileNets \cite{Howard2017mobile,mark2018mobile} employ depth-wise separable convolution to save computational budgets, where a standard convolution is decomposed into depth-wise and $1 \times 1$ point-wise convolution. ERFNet \cite{Romera2018erfnet} decomposes a 2D convolution (e.g., $3 \times 3$) into two 1D-factorized convolution (e.g., $3 \times 1$ and $1 \times 3$). An alternative efficient approach to lighten CNNs depends on group convolution \cite{Mehta2019espnet,krizhevsky2012imagenet}, where input channels and filter kernels are accordingly factored into a set of groups and each group is convolved independently. In spite of achieving impressive results, these lightweight networks prefer to adopt shallow network architectures to reduce model complexity, which may weaken the representation ability of visual data, leading to the degradation of performance.

In this paper, we introduce a novel lightweight network, called \emph{FDDWNet}, for the task of real-time semantic segmentation. Different from previous methods \cite{ma2018shufflenet,Mehta2019espnet,Paszke2016enet,Zhao2018ICnet,wu2018cgnet}, our \emph{FDDWNet} makes an effort to design more deeper network architecture to develop the ability of feature representation, while maintains very fewer model parameters to accelerate inference speed. In addition, \emph{FDDWNet} has multiple branches of skipped connections from intermediate convolution layers, in spite of adding a bit of computational burden, but helping to gather more context. The core unit of \emph{FDDWNet} is \underline{E}xtremely \underline{E}fficient \underline{R}esidual \underline{M}odule (EERM) using factorized dilated depth-wise separable convolutions (FDDWC), which allow us to learn feature representations from different scales of receptive fields with small number of model parameters. In summary, the contributions of this paper are three-folds: (1) More deeper architecture of \emph{FDDWNet} results in more powerful representation ability, thus yielding higher segmentation accuracy. (2) The EERM unit leverages identify mapping and FDDWC, facilitating model training without gradient vanish and explosion while remaining small model size. Additionally, all operations of EERM are differentiable, indicating that the entire \emph{FDDWNet} can be trained end-to-end. (3) The experiments show that \emph{FDDWNet} outperforms most lightweight networks in terms of available trade-off between speed and accuracy. \emph{FDDWNet} only has 0.8M model size, while achieves 60 FPS running speed on a single RTX 2080Ti GPU. On the other hand, \emph{FDDWNet} achieves a mean intersection over union (mIoU) of 71.5\% and 66.9\% on CityScapes \cite{Cordts2016the} and CamVid \cite{Brostow2008segmentation} datasets, respectively, without using any other postprocess or augmented training data.

\begin{table}[!b]
\tabcolsep 1.0mm \caption{Comparison between different types of convolutions. $n_r = (n - 1) \cdot r + 1$ is dilation rate, and $g$ indicates the number of groups. ``DW'' and ``DDW'' denote depth-wise separable convolution with or without dilation.}
\begin{center}
\begin{tabular}{l|c|c}
\toprule
\textbf{Convolutional type}                              &\textbf{Parameters}                 &\textbf{Size of receptive field} \\
\hline
Standard \cite{He2016deep}                               &$n^2c\hat{c}$                       &$n \times n$\\
Group \cite{krizhevsky2012imagenet}                      &$n^2c\hat{c}/g$                     &$n \times n$\\
1D-factorized \cite{Romera2018erfnet}                    &$2nc\hat{c}$                        &$n \times n$\\
DW \cite{ma2018shufflenet,mark2018mobile}                &$n^2c + c\hat{c}$                   &$n \times n$\\
DDW \cite{Mehta2019espnet}                               &$n^2c + c\hat{c}$                   &$n_r \times n_r$\\
our FDDWC                                                &$2nc + c\hat{c}$                    &$n_r \times n_r$\\
\bottomrule
\end{tabular}
\end{center}\label{tab:}
\end{table}

\section{Our Method}
\label{sec:FDDSNet}


\subsection{FDDWC}
\label{sec:EERM}

A standard convolution convolves an input feature tensor $\emph{\textbf{X}} \in \mathbb{R}^{W \times H \times c}$ with a filter kernel $\emph{\textbf{K}} \in \mathbb{R}^{n \times n \times c \times \hat{c}}$ to produce an output $\emph{\textbf{Y}} \in \mathbb{R}^{W \times H \times \hat{c}}$. It learns $n^2c\hat{c}$ model parameters with $n \times n$ receptive field. In contrast, FDDWC reduces computational burden by decomposing a standard convolution into two steps: (1) factorized depth-wise convolution with dilated rate $r$, performing independently lightweight filtering with two 1D-factorized filter kernels (e.g., $1 \times n$ and $n \times 1$) per input channel; and (2) a $1 \times 1$ point-wise convolution that learns a linear combinations of the input channels. This factorization reduces the model parameters from $n^2c\hat{c}$ to $2nc + c\hat{c}$. Tab. \ref{tab:} compares different types of convolutions, and demonstrates that our FDDWC is more efficient and can learn feature representations from different scales of receptive fields.

\begin{figure}[!t]
\centerline{\includegraphics[width = 0.5\textwidth]{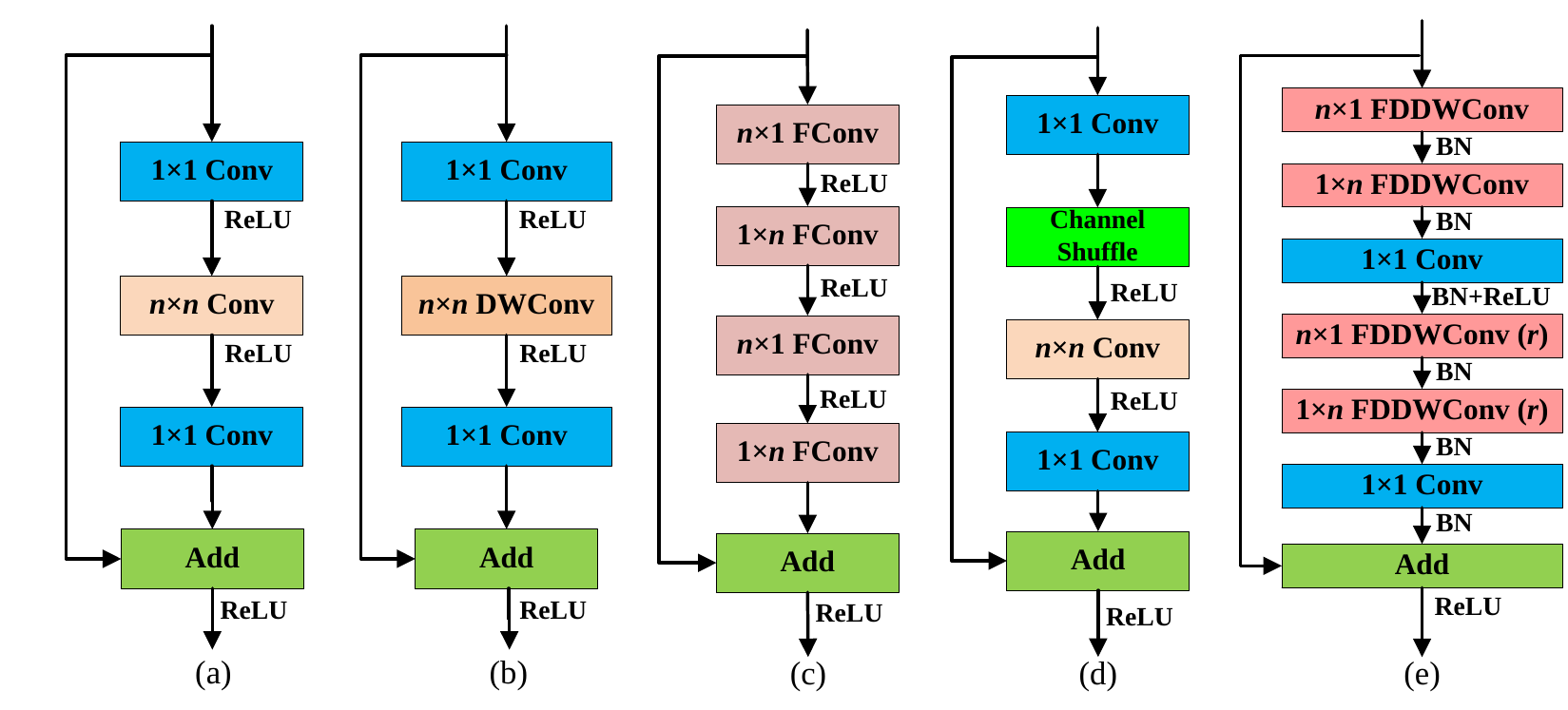}}
\caption{Comparison of different types of residual modules. From left to right are (a) bottleneck \cite{Paszke2016enet}, (b) MobileNet \cite{mark2018mobile}, (c) non-bottleneck-1D \cite{Romera2018erfnet}, (d) ShuffleNet \cite{zhang2018shuffle}, and (e) our EERM. ``Conv'' is a standard convolution, ``DWConv'' indicates depth-wise separable convolution, and ``FConv'' denotes 1D-factorized convolution. (Best viewed in color)} \label{fig:module}
\end{figure}

\subsection{EERM Unit}
\label{sec:EERM}

The recent years has witnessed many efficient residual modules, such as bottleneck (Fig. \ref{fig:module}(a)), MobileNet module (Fig. \ref{fig:module}(b)), non-bottleneck-1D (Fig. \ref{fig:module}(c)), and ShuffleNet module (Fig. \ref{fig:module}(d)). These widely-used residual modules, however, suffer from some limitations in terms of learning capacity and efficiency, while our aim here is to achieve the best possible trade-off between accuracy and efficiency. Towards this end, this section introduces EERM, the core unit of \emph{FDDWNet}, to approach the representational power of larger and denser layers, but at a considerably lower computational budgets.

EERM unit, which leverages the residual connections and FDDWC, combines the strength of 1D-factorized convolution \cite{Romera2018erfnet} and dilated depth-wise separable convolution \cite{Mehta2019espnet}. More specifically, as shown in Fig. \ref{fig:module}(e), a set of specialized 1D filter kernels (e.g., $1 \times n$ and $n \times 1$) are sequentially convolved with input per each channel, resulting in the independent filtering responses along output channels. Thereafter, a $1 \times 1$ point-wise convolution is used to recover channel dependency by learning a linear combinations of the input channels. Note these two operations are duplicated with dilated rate $r$ in each EERM unit. Finally, to facilitate training, the transferred output is added with input through the branch of identity mapping. Thanks to FDDWC, the total computational cost is reduced by a factor of $\frac{1}{n}(\frac{2}{\hat{c}}+ \frac{1}{n})$ with respect to a standard convolution. For example, EERM unit saves $8.6 \times$ floating point operations (FLOPs) when $n = 3$ (adopted in our \emph{FDDWNet}) and $\hat{c} = 128$, leading to the higher efficiency, especially when the entire network has more deeper architecture. We also use different dilated rate $r$ to enlarge receptive fields and at the same time remove the gridding artifacts. Although the focus of this paper is semantic segmentation, we believe that EERM can be easily transferred to any existing network architectures that are used for other visual tasks, such as object detection \cite{Girshick2015fast,redmon2016you} and image classification \cite{krizhevsky2012imagenet,He2016deep,going2015szegedy}.

\begin{table}[!t]
\small \tabcolsep 1.5mm \caption{The architecture of \emph{FDDWNet}. ``Size'' denotes the dimension of output feature maps, $C$ is the number of classes.}
\begin{center}
\begin{tabular}{c|c|l|c}
\toprule
\textbf{Stage} & \textbf{Layer} & \textbf{Type} &\textbf{Size} \\
\midrule
\multirow{13}*{\rotatebox{90} {Encoder}}
&1     &\textbf{Downsampling Unit}          &$512 \times 256 \times 16$ \\		
&2     &\textbf{Downsampling Unit}          &$256 \times 128 \times 64$ \\
&3-7   &$5 \times$ \textbf{EERM} ($r = 1$)  &$256 \times 128 \times 64$ \\
&8     &\textbf{Downsampling Unit}          &$128 \times 64 \times 128$ \\ \cmidrule{2-4}
&\multirow{1}{*}{9-16}     &$ \left [ \begin{array}{c} \textbf{EERM}~(r = 1)\\ \textbf{EERM}~(r = 2)\\
                                                        \textbf{EERM}~(r = 5)\\ \textbf{EERM}~(r = 9)
                              \end{array}  \right ] \times 2 $     &$128 \times 64 \times 128$ \\ \cmidrule{2-4}
&\multirow{1}{*}{17-24}    &$ \left [ \begin{array}{c} \textbf{EERM}~(r = 2)\\ \textbf{EERM}~(r = 5)\\
                                                       \textbf{EERM}~(r = 9)\\ \textbf{EERM}~(r = 17)
                              \end{array}  \right ] \times 2 $     &$128 \times 64 \times 128$ \\ \midrule
\multirow{5}*{\rotatebox{90} {Decoder}}
&25   &\textbf{Upsampling Unit}            &$256 \times 128 \times 64$ \\
&26-27&$2 \times$ \textbf{EERM} ($r = 1$)  &$256 \times 128 \times 64$ \\
&28   &\textbf{Upsampling Unit}            &$512 \times 256 \times 16$ \\
&29-30&$2 \times$ \textbf{EERM} ($r = 1$)  &$512 \times 256 \times 16$ \\
&31   &\textbf{Upsampling Unit}            &$1024 \times 512 \times C$ \\
\bottomrule
\end{tabular}
\end{center}\label{tab:FDDSNet}
\end{table}

\subsection{Network Architecture}
\label{sec:Architecture}

Our \emph{FDDWNet} is built mainly based on EERM unit, which enables us to explore more deeper architecture, but with very smaller computational overhead. \emph{FDDWNet} follows an EDN-like structure \cite{Romera2018erfnet,Paszke2016enet}, where the encoder produces the downsampling features, and the decoder subsequently upsamples these features to match input resolution. In addition, skipped connections \cite{long2017fully,zhao2017pyramid} between the encoder and the decoder are also used to improve segmenting accuracy. The main network architecture is depicted in Tab. \ref{tab:FDDSNet}. The encoder is consist of layers from 1 to 24, including EERM and downsampling units. Downsampling enables more deeper network to gather context, while at the same time helps to reduce computation. On the other hand, layers from 25 to 31 form the decoder, composed of EERM and upsampling units. To build skipped connections, layer 7 undergoes two extra EERMs, and then added to two upsampling units in decoder. Note before added to layer 28, layer 7 has to be $2\times$ upsampled, resulting in the equal resolutions for following convolution and fusion.

\section{Experiments}
\label{sec:experiments}

\subsection{Implementation Details}\label{sec:Implementation}

\textbf{Dataset.} We test \emph{FDDWNet} on Cityscapes \cite{Cordts2016the} and CamVid \cite{Brostow2008segmentation} datasets, which are widely-used benchmarks for the task of real-time semantic segmentation. Both datasets focus on street scenes segmentation for self-driving, where the first one includes 19, and the second one involves 11 object categories. Specifically, the Cityscapes dataset contains 5,000 pixel-wise annotated images and 20,000 coarsely annotated images with $2048 \times 1024$ image resolution collected from 50 cities. Following \cite{Romera2018erfnet}, we only employ images with fine annotations, resulting in 2,975 training, 500 validation and 1,525 testing images. Camvid, on the other hand, is a smaller dataset, collected from 5 videos. It only includes 367 training, 101 validation, and 233 testing images with resolution $960 \times 720$.

\begin{table}[!t]
\small \tabcolsep 1.8mm \caption{Comparison with the state-of-the-art approaches in terms of accuracy and efficiency. `-/-' represents the results on test set of CityScapes and CamVid datasets, respectively.}
\begin{center}
\begin{tabular}{c|cccc}
\toprule
Method  & Layers & Cla (\%) &Speed(FPS) &Para(M) \\
\midrule
CGNet \cite{wu2018cgnet}                 &28    &64.8/65.6     &17/~~- &\textbf{0.50} \\
ESPNet V2 \cite{Mehta2019espnet}        &-     &66.2/~~-~~~   &67/~~-~ &1.25 \\
ERFNet \cite{Romera2018erfnet}          &23    &68.0/~~-~~~   &42/~~- &2.1 \\
FSCNNet \cite{Poudel2019fast}           &19    &68.0/~~-~~~   &\textbf{124}/~~-~  &1.11       \\
DSNet \cite{wang2018dsnet}              &28    &69.3/~~-~~~   &37/~~-  &0.91       \\
ICNet \cite{Zhao2018ICnet}              &-     &69.5/\textbf{67.1}     &30/28 &26.5 \\
DABNet \cite{li2019dabnet}              &16    &70.1/66.4     &104/\textbf{146}  &0.76       \\
\midrule
Ours                                    &\textbf{31}    &\textbf{71.5}/66.9 &60/79 &0.80 \\
\bottomrule
\end{tabular}
\end{center}\label{tab:Result}
\end{table}

\begin{figure}[!t]
\centerline{\includegraphics[width = 0.5\textwidth]{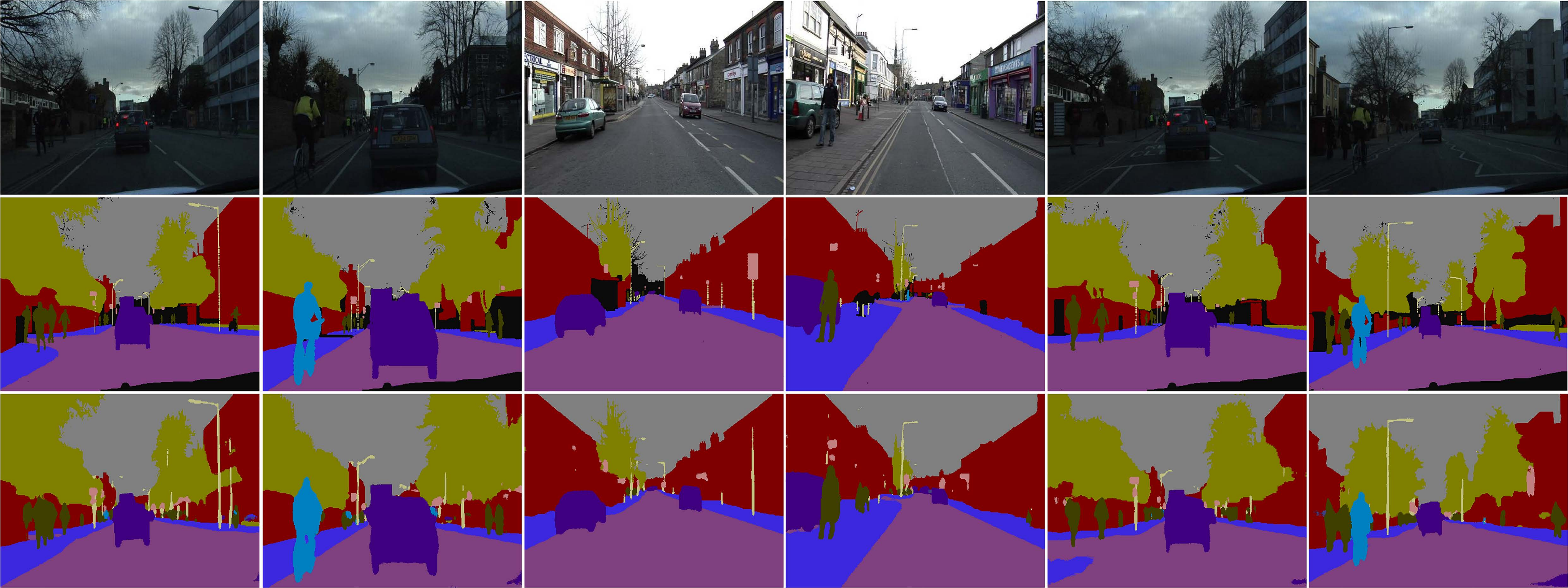}}
\caption{The visual examples of our method on CamVid test set. From top to bottom are original images, ground truth, and our segmentation results. (Best viewed in color)} \label{fig:CamVid}
\end{figure}

\noindent \textbf{Baselines.} To show the advantages of \emph{FDDWNet}, we selected 7 state-of-the-art lightweight networks as baselines, including ERFNet \cite{Romera2018erfnet}, ESPNet V2 \cite{Mehta2019espnet}, ICNet \cite{Zhao2018ICnet}, CGNet \cite{wu2018cgnet}, FSCNNet \cite{Poudel2019fast}, DSNet \cite{wang2018dsnet}, and DABNet \cite{li2019dabnet}.

\noindent \textbf{Parameter settings.} \emph{FDDWNet} is implemented on the hardware platform of Dell workstation with a single RTX 2080Ti GPU, and trained in an end-to-end manner using stochastic gradient descent algorithm \cite{Bottou2010sgd}. We favor a large minibatch size (set as 4) to make full use of the GPU memory, where the initial learning rate is $10^{-3}$ and the `poly' learning rate policy \cite{zhao2017pyramid} is adopted with power 0.9, together with momentum and weight decay are set to 0.9 and $10^{-5}$, respectively.

\begin{table*}[!t]
\small \tabcolsep 1.1mm \caption{Individual category results on the CityScapes {test} set in terms of class mIoU scores.}
\begin{center}
\begin{tabular}{c|ccccccccccccccccccc|c}
\toprule
Method &{Roa}  &{Sid}  &{Bui}  &{Wal}  &{Fen}  &{Pol}  &{TLi}  &{TSi}  &{Veg}  &{Ter}  &{Sky}  &{Ped}  &{Rid}  &{Car}  &{Tru}  &{Bus}  &{Tra}  &{Mot}  &{Bic}  &{Cla}  \\
\midrule
CGNet \cite{wu2018cgnet} &95.5 &78.7 &88.1 &40.0 &43.0 &54.1 &59.8 &63.9 &89.6 &67.6 &92.9 &74.9 &54.9 &90.2 &44.1 &59.5 &25.2 &47.3 &60.2 &64.8 \\
ESPNet V2 \cite{Mehta2019espnet} &97.3 &78.6 &88.8 &43.5 &42.1 &49.3 &52.6 &60.0 &90.5 &66.8 &93.3 &72.9 &53.1 &91.8 &53.0 &65.9 &53.2 &44.2 &59.9 &66.2 \\
FSCNNet \cite{Poudel2019fast} &97.9 &81.6 &89.7 &46.4 &48.6 &48.3 &53.0 &60.5 &90.7 &67.2 &94.3 &74.0 &54.6 &93.0 &\textbf{57.4} &65.5 &\textbf{58.2} &50.0 &61.2 &68.0 \\
ERFNet \cite{Romera2018erfnet}&97.7 &81.0 &89.8 &42.5 &48.0 &56.3 &59.8 &65.3 &91.4 &68.2 &94.2 &76.8 &57.1 &92.8 &50.8 &60.1 &51.8 &47.3 &61.7 &68.0 \\
DSNet \cite{wang2018dsnet}&97.1 &79.7 &89.4 &37.8 &50.4 &56.7 &63.0 &68.5 &91.0 &67.9 &93.5 &75.7 &\textbf{61.4} &91.9 &50.4 &66.8 &56.8 &54.0 &65.4 &69.3 \\
ICNet \cite{Zhao2018ICnet} &97.1 &79.2 &89.7 &43.2 &48.9 &\textbf{61.5} &60.4 &63.4 &91.5 &68.3 &93.5 &74.6 &56.1 &92.6 &51.3 &\textbf{72.7} &51.3 &53.6 &\textbf{70.5} &69.5 \\
DABNet \cite{li2019dabnet} &97.9 &82.0 &90.6 &45.5 &50.1 & 59.3 &63.5 & 67.7 &91.8 &70.1 &92.8 &78.1 &57.8 &93.7 &52.8 &63.7 &56.0 &51.3 &66.8 &70.1 \\
\midrule
Ours &\textbf{98.0} &\textbf{82.4} &\textbf{91.1} &\textbf{52.5} &\textbf{51.2} &59.9 &\textbf{64.4} &\textbf{68.9} &\textbf{92.5} &\textbf{70.3} &\textbf{94.4} &\textbf{80.8} &59.8 &\textbf{94.0} &56.5 &68.9 &48.6 &\textbf{55.7} &67.7 &\textbf{71.5} \\
\bottomrule
\end{tabular}
\end{center}\label{tab:CityScapes}
\end{table*}

\begin{figure*}[!t]
\centerline{\includegraphics[width = 1.0\textwidth]{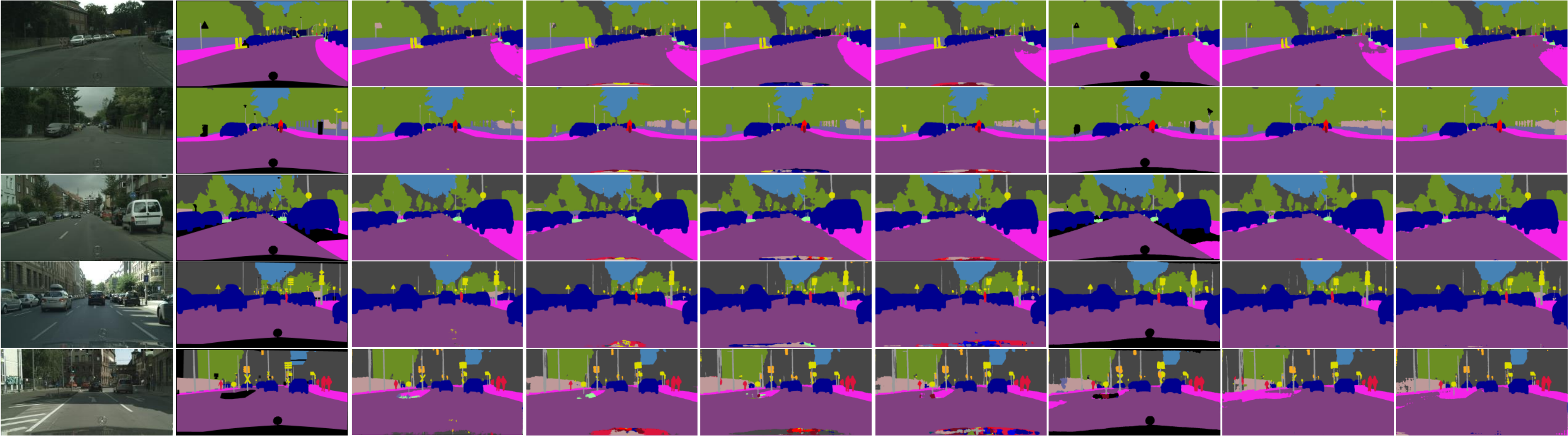}}
\caption{The visual comparison on Cityscapes val set. From left to right are images, ground truth, segmentation outputs from \emph{FDDWNet}, DABNet \cite{li2019dabnet}, DSNet \cite{wang2018dsnet}, FSCNNet \cite{Poudel2019fast}, ESPNet V2 \cite{Mehta2019espnet}, ERFNet \cite{Romera2018erfnet} and CGNet \cite{wu2018cgnet}. (Best viewed in color)} \label{fig:CityScapes}
\end{figure*}

\subsection{Evaluation Results on Cityscapes}\label{sec:CityscapesResults}

Tab. \ref{tab:Result} and Tab. \ref{tab:CityScapes} compare our \emph{FDDWNet} with selected state-of-the-art networks. The results show that \emph{FDDWNet} outperforms these baselines in terms of segmentation accuracy and implementing efficiency. In spite of having deepest network architecture, our method only has 0.8M model size, while achieves 60 FPS inference speed and 71.5\% mIoU without using extra training data. From Tab. \ref{tab:CityScapes}, it is observed that 13 out of the 19 object categories obtains best mIoU scores, especially for some categories, achieving remarkable improvement (e.g., 6.1\% for `Wal' and 2.7\% for `Ped'). Regarding to the efficiency, \emph{FDDWNet} is nearly $2 \times$ faster and $53 \times$ smaller than ICNet \cite{Zhao2018ICnet}. Although FSCNNet \cite{Poudel2019fast}, another efficient network, is nearly $2 \times$ efficient, but has $1.4 \times$ larger model size and delivers poor segmentation accuracy of 3.5\% drop than our \emph{FDDWNet}. Another intriguing result is that, due to the lightweight design of EERM unit, our approach has comparable model size, but nearly $2 \times$ network depth with DABNet \cite{li2019dabnet}. Fig. \ref{fig:CityScapes} shows some qualitative results on the CityScapes val set. It is demonstrated that, compared with baselines, \emph{FDDWNet} not only correctly classifies object instances, but also produces consistent visual outputs.

\subsection{Evaluation Results on CamVid}\label{sec:CamVidResults}

We also evaluate \emph{FDDWNet} on CamVid \cite{Brostow2008segmentation} dataset, and report the results on Tab. \ref{tab:Result}. Compared with the selected state-of-the-art baselines, \emph{FDDWNet} also shows the superior performance in terms of running speed and segmenting accuracy. Compared with ICNet \cite{Zhao2018ICnet}, \emph{FDDWNet} achieves slightly performance drop (0.2\% of mIoU), but runs $2.8 \times$ faster. Note \emph{FDDWNet} performs faster on CamVid dataset  (79 vs. 60 FPS), due to its smaller input image resolutions ($2048 \times 1024$ of Cityscapes and $960 \times 720$ of CamVid). Several visual examples of segmentation outputs are shown in Fig. \ref{fig:CamVid}.

\section{Conclusion Remarks and Future Work}
\label{sec:conclusion}

This paper has presented a \emph{FDDWNet}, which explores a more deeper EDN-like architecture for real-time semantic segmentation. \emph{FDDWNet} employs EERM to leverage model training and running speed. To improve efficiency, a factorized dilated depth-wise separable convolution is used in EERM, allowing us to approach more powerful representation with different scales of receptive fields, but at a considerably lower computational budgets. The experimental results show that our \emph{FDDWNet} achieves best trade-off on CityScapes and CamVid datasets in terms of segmentation accuracy and implementing efficiency. In the future, we are interested in transferring EERM to the visual tasks of object detection \cite{Girshick2015fast,redmon2016you} and image classification \cite{krizhevsky2012imagenet,He2016deep,going2015szegedy}, still resulting in lightweight networks while remaining high accuracy.

\bibliographystyle{IEEEbib}
\bibliography{strings,refs}

\end{document}